%% file: Reyes.Abad.ISERC.16.tex
\documentclass{iserc}
\usepackage[ vlined, linesnumbered, ruled]{algorithm2e}
\usepackage{ multirow}
\usepackage{ epstopdf}

\newcommand{\eg}{\textit{e.g.,\;}}

\newcommand{\ie}{\textit{i.e.,\;}}

\newcommand{\Ie}{\textit{I.e.,\;}}
\newcommand{\Iec}{\textit{I.e.:\;}}
\newcommand{\fromto}[2]{ \{ \, #1, \, \dots, \, #2 \} }
\newcommand{\define}{\triangleq}

\DeclareMathOperator*{\argmax}{arg\,max}

\renewcommand{\Pr}{\mathbb{P}} 
\renewcommand{\Re}{\mathbb{R}} 
\newcommand{\ReSP}{{\Re}_{> 0}} 

\newcommand{\ZNN}{\mathbb{Z}_{\geq 0}} 

\conference{Proceedings of the 2016 Industrial and Systems Engineering Research Conference\\
H. Yang, Z. Kong, and MD Sarder, eds.} 
\title{A Dynamic Bayesian Network Model for \\ Inventory Level Estimation in Retail Marketing}
\author{ Luis I. Reyes-Castro \qquad Andres G. Abad \\
Escuela Superior Polit\'ecnica del Litoral (ESPOL) \\ Guayaquil - Ecuador } 
\authorlist{Reyes-Castro and Abad} 
\abstractID{1028} 

\begin{document}
\allowdisplaybreaks
\maketitle

\begin{abstract}
Many retailers today employ inventory management systems based on Re-Order Point Policies, most of which rely on the assumption that all decreases in product inventory levels result from product sales. Unfortunately, it usually happens that small but random quantities of the product get lost, stolen or broken without record as time passes, \eg as a consequence of shoplifting. This is usual for retailers handling large varieties of inexpensive products, \eg \linebreak grocery stores. In turn, over time these discrepancies lead to \emph{stock freezing} problems (see Ref. \cite{Kang.IEET2005information}), \ie situations where the system believes the stock is above the re-order point but the actual stock is at zero, and so no replenishments or sales occur. Motivated by these issues, we model the interaction between sales, losses, replenishments and inventory levels as a Dynamic Bayesian Network (DBN), where the inventory levels are unobserved (\ie hidden) variables we wish to estimate. We present an Expectation-Maximization (EM) algorithm to estimate the parameters of the sale and loss distributions, which relies on solving a one-dimensional dynamic program for the E-step and on solving two separate one-dimensional nonlinear programs for the M-step. 
\end{abstract}

\section*{Keywords}
\vspace{0.5\baselineskip}
Inventory Management, Inventory Record Inaccuracies, Inventory Shrinkage, Dynamic Bayesian Networks. 

\section{Introduction} \label{sec:Intro}
\vspace{0.5\baselineskip}

We consider the sale and replenishment of a product in a store over a time horizon of $T \geq 2$ time periods. We let $I_0$ denote the initial inventory level, and for each period $t \in \fromto{1}{T}$ we let $I_t$ denote the inventory level at the end of that period. Furthermore, for each period $t$ we let $S_t$ denote the random but observed number of units of the product sold during that period. Moreover, we assume that replenishments happen after all the sales of the period have been completed, \eg after the store closes for the day, and for each period $t$ we let $R_t$ denote the non-random and observed number of units of the product replenished at the end of that period. Finally, and most importantly, we assume that on each time period some number of units of the product may be lost, broken or stolen without knowledge of the store's manager, \ie without record. In particular, for each period $t$ we let $L_t$ denote the random and unobserved number of units of the product lost, broken or stolen during that period. For modeling reasons, we further assume that in each period all losses occur after all sales have been completed but before any replenishments arrive, although in reality sale, loss and replenishment epochs may intertwine. 

Since the product losses are unobserved, so are the inventory levels, which motivates the main problem of this paper: \linebreak Estimating the (unknown) sale and loss distribution parameters along with the (unobserved) inventory levels. \linebreak This problem is important because having a good model of the inventory level history of a product is essencial to knowing when to re-order it so as to keep it available to the customers. Unfortunately, most studies so far have focused \linebreak on qualitatively describing the problem and on proposing heuristic replenishment and inspection policies; the reader is refered to Refs. \cite{Kang.IEET2005information,Iglehart.Morey.MS.72,Raman.DeHoratius.CMR.01,DeHoratius.Raman.MS.08,DeHoratius.MSOM.08,Chen.Mersereau.RSCM.15} for a sample of previous work. For instance, Ref. \cite{DeHoratius.MSOM.08}, which in our opinion is the study most closely related to our paper, describes a method for estimating the aforementioned parameters by collecting statistics of past inventory inspection data and pooling the statistics associated with similar products. In contrast, we propose an algorithm capable of estimating the parameters even in the absence of past inspection data. 

\section{Assumptions, Problem Statement, and Solution Method} \label{sec:Probabilistic_Assumptions}
\vspace{0.5\baselineskip}

As usual, we assume that for each time period the physical inventory level at the end of the period is equal to the physical inventory level at the end of the previous period, minus the product sales and losses during the period, \linebreak plus the replenishments at the end of the period. \Iec 
\vspace{0.20\baselineskip}
\begin{equation}
\label{eq:Inventory_Level}
\forall \, t \in \fromto{1}{T} \; \colon \; 
I_t \, = \, I_{t-1} - S_t - L_t + R_t
\end{equation}

Furthermore, for each period $t$ we assume that $S_t$ has a truncated Poisson distribution with parameter $\sigma > 0$ and upper bound $I_{t-1}$. \Ie if $(X_t)_{1 \leq t \leq T}$ is a sequence of i.i.d. Poisson random variables with parameter $\sigma$ then for each $t$ we have $S_t = \min \, \{ \, X_t, \, I_{t-1} \, \}$. We choose the Poisson distribution because it is commonly used to model random demands, although it is fairly straighforward to extend our model to one with a different sales distribution, \eg \linebreak Bernoulli, Geometric, Binomial, etc. The truncation is justified because in each period the number of units of the product that the store can sell is limited by the product's physical inventory level at the end of the previous period; \linebreak we do not allow backordering. Moreover, since the value of $I_{t-1}$ is all we need to describe the distribution of $S_t$, \linebreak we observe that conditional on $I_{t-1}$ the random variable $S_t$ is independent of all inventory levels up to period $t-2$ and of all sales, losses and replenishments up to period $t-1$. More precisely, for each $t \in \fromto{1}{T}$ : 
\vspace{0.20\baselineskip}
\begin{equation}
\label{eq:Conditional_Independecies_of_St}
\Pr \, ( \, S_t \mid (I_{\tau})_{0 \leq \tau \leq t-1}, \, (S_{\tau})_{1 \leq \tau \leq t-1}, \, (L_{\tau})_{1 \leq \tau \leq t-1}, \, (R_{\tau})_{1 \leq \tau \leq t-1} \, ) \; = \; 
\Pr \, ( \, S_t \mid I_{t-1} \, ) 
\end{equation}

In addition, for each period $t$ we assume that $L_t$ has a truncated Bernoulli distribution with parameter $\lambda \in [0,1]$ and upper bound $I_{t-1} - S_t$, \ie if $(Y_t)_{1 \leq t \leq T}$ is a sequence of i.i.d. Bernoulli random variables with parameter $\lambda$ then for each $t$ we have $L_t = \min \, \{ \, Y_t, \, I_{t-1} - S_t \, \}$. We choose the Bernoulli distribution because we are interested in modeling small loss rates (\ie rates of no more than a unit per period), but our model also allows other discrete distributions. \linebreak The truncation relies on the fact that in each period all losses occur once all sales have been completed but before any replenishments. Also, since the values of $I_{t-1}$ and $S_t$ completely specify the distribution of $L_t$, we see that conditional on $I_{t-1}$ and $S_t$ the random variable $L_t$ is independent of all inventory levels up to period $t-2$ and of all sales, losses and replenishments up to period $t-1$. More precisely, for each $t \in \fromto{1}{T}$ : 
\vspace{0.20\baselineskip}
\begin{equation}
\label{eq:Conditional_Independecies_of_Lt}
\Pr \, ( \, L_t \mid (I_{\tau})_{0 \leq \tau \leq t-1}, \, (S_{\tau})_{1 \leq \tau \leq t-1}, \, (L_{\tau})_{1 \leq \tau \leq t-1}, \, (R_{\tau})_{1 \leq \tau \leq t-1} \, ) \; = \; 
\Pr \, ( \, L_t \mid I_{t-1}, \, S_t \, )
\end{equation}

Now, we can continue building our model without product loss variables (\ie $L_t$'s) if we note, from Equation (\ref{eq:Inventory_Level}), \linebreak that for each $t \in \fromto{1}{T}$ : 
\vspace{0.20\baselineskip}
\begin{equation}
\label{eq:Conditional_Distribution_of_It}
\Pr \, ( \, I_t \mid I_{t-1}, \, S_t, \, R_t \, ) 
\; = \; \Pr \, ( \, L_t = I_{t-1} - S_t + R_t - I_t \mid I_{t-1}, \, S_t \, )
\end{equation}

Furthermore, combining Equations (\ref{eq:Conditional_Independecies_of_Lt}) and (\ref{eq:Conditional_Distribution_of_It}), we see that conditional on $I_{t-1}$, $S_t$ and $R_t$ the random variable $I_t$ is independent of all inventory levels up to period $t-2$ and of all sales, losses and replenishments up to period $t-1$. \linebreak Moreover, since our assumptions require that for each period $t$ all sales and losses occur before any replenishments, we see that $S_t + L_t \leq I_{t-1}$, and so replacing $I_{t-1}$ using Equation (\ref{eq:Inventory_Level}) we obtain the bound $R_t \, \leq \, I_t$. In turn, combining this bound with the same equation and the fact that $L_t \in \{ \, 0, 1 \}$ surely, we have: 
\vspace{0.20\baselineskip}
\begin{equation}
\forall \, t \in \fromto{1}{T} \; \colon \; 
\max \, \{ \, R_t \, , \, I_{t-1} - S_t + R_t - 1 \, \} \, \leq \, 
I_t \, \leq \, I_{t-1} - S_t + R_t
\label{eq:It_LB_UB}
\end{equation}

Finally, the problem we seek to solve is that of finding a Maximum Likelihood Estimate (MLE) of the sale and loss distribution parameters $(\sigma,\lambda)$, respectively, and of the unobserved inventory level history $I \define (I_t)_{1 \leq t \leq T}$, given an initial inventory level $I_0$, a sales history $S \define (S_t)_{1 \leq t \leq T}$, and a replenishment history $R \define (R_t)_{1 \leq t \leq T}$. More precisely, we seek to solve the following Mixed-Integer Nonlinear Program (MINLP): 
\vspace{0.20\baselineskip}
\begin{align}
\text{maximize:} \quad 
& \log \, \Pr_{ \sigma, \lambda} \, ( \, I \mid I_0, \, S, \, R \, ) \label{eq:OMP_utility_function} \\
\text{subject to the constraints:} \quad
& \sigma > 0; \; \lambda \in [0,1]; \; I \in \ZNN^T; \; 
\text{Inequality (\ref{eq:It_LB_UB})} \label{eq:OMP_constraints}
\end{align}

Unfortunately, to the best of our knowledge there are no efficient methods for computing a global maximizer to \linebreak Problem (\ref{eq:OMP_utility_function})-(\ref{eq:OMP_constraints}) with provable guarantees. Therefore, in this work we present an Expectation-Maximization (EM) \linebreak Algorithm, which is a greedy algorithm, to compute a local maximizer of the aforementioned function (see Ref. \cite{Bishop.06}). \linebreak The algorithm itself is quite simple, relying on the following iteration: 
\begin{enumerate}
\item Select initial guesses for the MLEs of the sale and loss distribution parameters, denoted $( \, \sigma^{[0]}, \, \lambda^{[0]} )$. 
\item For each iteration $k \geq 1$ : 
\begin{enumerate}
\item \textbf{E-step:} Using the previous iteration's MLEs of the sale and loss parameters, \ie $( \, \sigma^{[k-1]}, \, \lambda^{[k-1]} )$, \linebreak compute the current MLE of the inventory level history, denoted $I^{[k]} \define (I_t)_{1 \leq t \leq T}^{[k]}$. 
\item \textbf{M-step:} Using the current MLE of the inventory level history, \ie $I^{[k]}$, compute the current MLEs of the sale and loss parameters, denoted $( \sigma^{[k]}, \, \lambda^{[k]} )$. 
\item If the sale and loss parameters haven't changed, \ie if $( \, \sigma^{[k-1]}, \, \lambda^{[k-1]} ) = ( \sigma^{[k]}, \, \lambda^{[k]} )$, then terminate. 
\end{enumerate}
\end{enumerate}
Execution of the E-step and of the M-step is explained in the following two sections. 

\section{E-step: Estimation of the Inventory Level History} \label{sec:E-step}
\vspace{0.5\baselineskip}

In this section we seek to compute an inventory level history $I \in \ZNN^T$ of maximum log-likelihood, among all those which satisfy Inequality (\ref{eq:It_LB_UB}), given the sale and loss distribution parameters $\sigma \in \Re_{>0}$ and $\lambda \in [0,1]$ and conditional \linebreak on the initial inventory level $I_0 \in \ZNN$, the sales history $S \in \ZNN^T$, and the replenishment history $R \in \ZNN^T$. \linebreak To start, we note that since $I_0$, $S$ and $R$ are fixed, so is their joint likelihood, \ie $\Pr_{ \sigma, \lambda} ( \, I_0, \, S, \, R \, )$. Therefore, \linebreak instead of seeking an inventory level history of maximum conditional log-likelihood, we can search for one that maximizes the joint log-likelihood of all the variables, \ie the function $I \mapsto \log \, \Pr_{ \sigma, \lambda} ( \, I_0, \, I, \, S, \, R \, )$. 

Next, we recognize that the joint likelihood of all the variables can be factored as a Dynamic Bayesian Network (DBN), \ie a Bayesian Network with a chain-like structure where each type of node represents a time history (see Ref. \citep{Koller.Friedman.09}). Certainly, from the statistical assumptions put forth in Section \ref{sec:Probabilistic_Assumptions}, we can represent the joint likelihood function as a DBN where each variable is a node, and where the nodes' parents are as follows: 
\begin{itemize}
\item Nodes $I_0$, $R_1$, $\dots$, and $R_T$ have no parents. 
\item For each period $t \in \fromto{1}{T}$, the parent of node $S_t$ is node $I_{t-1}$. 
\item For each period $t \in \fromto{1}{T}$, the parents of node $I_t$ are nodes $I_{t-1}$, $S_t$ and $R_t$. 
\end{itemize}

\begin{figure}[htb]
\centering
\small
\def\svgwidth{0.85\columnwidth}
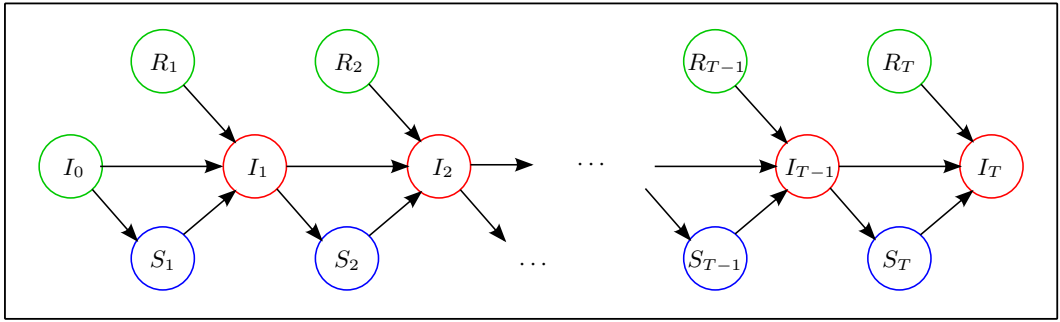
\caption{Our Dynamic Bayesian Network (DBN) model. Here, deterministic variable nodes are shown in green, observed random variable nodes are shown in blue, and unobserved random variable nodes are shown in red.}
\end{figure}

Now, with this representation in mind we can factor the joint log-likelihood function as follows: 
\vspace{0.20\baselineskip}
\begin{align}
\log \, \Pr_{ \sigma, \lambda} ( \, I_0, \, I, \, S, \, R \, ) \; 
& = \; \log \left( \;
\prod_{t=1}^T \, \Pr_{\sigma}( \, S_{t} \mid I_{t-1} \, ) \, 
\Pr_{\lambda}( \, I_t \mid I_{t-1}, \, S_t, \, R_t \, ) \, \right) \notag \\
& = \; \sum_{t=1}^{T-1} \bigg( \, 
\log \, \Pr_{\sigma}( \, S_{t} \mid I_{t-1} \, ) + 
\log \, \Pr_{\lambda}( \, I_t \mid I_{t-1}, \, S_t, \, R_t \, ) \, \bigg) 
\label{eq:JLL_DBN_factorization}
\end{align}

Furthermore, if for each $t \in \{ \, 1, \, \dots, \, T \}$ we define the function 
\vspace{0.20\baselineskip}
\begin{equation}
\phi_{ \sigma, \lambda}^{[t]}( \, I_{t-1}, \, I_{t} \, ) \; \define \; 
\log \, \Pr_{\sigma}( \, S_{t} \mid I_{t-1} \, ) + 
\log \, \Pr_{\lambda}( \, I_t \mid I_{t-1}, \, S_t, \, R_t \, ) \, ,
\label{eq:phi_t}
\end{equation}

then it is clear that our problem is equivalent to that of maximizing the function: 
\vspace{0.20\baselineskip}
\begin{equation}
\Phi_{ \sigma, \lambda}( \, I \, ) \; = \; 
\sum_{t=1}^{T} \phi_{ \sigma, \lambda}^{[t]}( \, I_{t-1}, \, I_{t} \, )
\label{eq:function_Phi}
\end{equation}

Moreover, the function above can be maximized sequentially by means of a Dynamic Programming (DP) approach. Indeed, notice the following: 
\vspace{0.20\baselineskip}
\begin{align}
& \max_{ I \, \in \, \ZNN^T } \Phi_{ \sigma, \lambda}( \, I \, ) \; 
= \; \max_{ I_1, \, \dots, \, I_{T-1} } 
\; \bigg[ \, \left( \; 
\sum_{t=1}^{T-1} \phi_{ \sigma, \lambda}^{[t]}( \, I_{t-1}, \, I_{t} \, ) \, \right) + \underbrace{ \, \max_{ I_T \, \in \, \ZNN } \; \phi_{ \sigma, \lambda}^{[T]}( \, I_{T-1}, \, I_T \, ) }_{ \displaystyle \define \; \psi_{ \sigma, \lambda}^{[T-1]}( \, I_{T-1} \,) } \, \bigg] 
\notag \\[-0.8ex]
& = \; \max_{ I_1, \, \dots, \, I_{T-2} } \; \bigg[ \, \left( \; 
\sum_{t=1}^{T-2} \phi_{ \sigma, \lambda}^{[t]}( \, I_{t-1}, \, I_{t} \, ) \, \right) + \underbrace{ \, \max_{ I_{T-1} \, \in \, \ZNN } \, \bigg[ \, \phi_{ \sigma, \lambda}^{[T-1]}( \, I_{T-2}, \, I_{T-1} \, ) + \psi_{ \sigma, \lambda}^{[T-1]}( \, I_{T-1} \,) \, \bigg] \, }_{ 
\displaystyle \define \; \psi_{ \sigma, \lambda}^{[T-2]}( \, I_{T-2} \,) } \, \bigg] \notag \\[-0.2ex]
& = \; \cdots \notag \\[-0.2ex]
& = \; \max_{ I_1, \, \dots, \, I_{T-k} } \; \bigg[ \, \left( \; 
\sum_{t=1}^{T-k} \phi_{ \sigma, \lambda}^{[t]}( \, I_{t-1}, \, I_{t} \, ) \, \right) + \underbrace{ \, \max_{ I_{T-k+1} \, \in \, \ZNN } \, \bigg[ \, \phi_{ \sigma, \lambda}^{[T-k+1]}( \, I_{T-k}, \, I_{T-k+1} \, ) + \psi_{ \sigma, \lambda}^{[T-k+1]}( \, I_{T-k+1} \,) \, \bigg] \, }_{ 
\displaystyle \define \; \psi_{ \sigma, \lambda}^{[T-k]}( \, I_{T-k} \,) } \, \bigg]
\label{eq:max_JLL_as_DP}
\end{align}

Hence, in light of the previous arguments, Algorithm \ref{algo:DP_for_the_E-step} computes a sequence of feasible inventory levels $I^* \in \ZNN^T$ which maximizes function $\Phi_{ \sigma, \lambda}$ among all feasible inventory level histories. 

\begin{algorithm}[htb]
\caption{Dynamic Programming Algorithm for the E-step}
\label{algo:DP_for_the_E-step}
\KwData{Sale and loss distribution parameters $\sigma \in \Re_{>0}$ and $\lambda \in [0,1]$, initial inventory level $I_0 \in \ZNN$, \newline sales history $S \define (S_t)_{1 \leq t \leq T} \in \ZNN^T$, replenishment history $R \define (R_t)_{1 \leq t \leq T} \in \ZNN^T$. }
\KwResult{Maximum likelihood estimate (MLE) of the inventory level history $I^* \define (I_t^*)_{1 \leq t \leq T}$.}
\tcp{SETUP: Computes upper bounds on the feasible inventory levels}
let $I_0^{max} = I_0$ \\
\ForEach{ $t \in ( \, 1, \, \dots, \, T \, )$ }{let $I_t^{max} = I_{t-1}^{max} - S_t + R_t$}
\tcp{BACKWARD PASS: Computes the optimal $I_{t+1}$'s as functions of the $I_{t}$'s}
let $\psi_{ \sigma, \lambda}^{[T]}$ be a zero function, \emph{i.e.,} $\psi_{ \sigma, \lambda}^{[T]} \; \colon \ZNN \mapsto \{0\}$ \\
\ForEach{ $t \in ( \, T-1, \, \dots, \, 0 \, )$ }
{
\ForEach{ $I_t \in \{ \, 0, \, \dots, \, I_t^{max} \, \}$ }
{
let $\psi_{ \sigma, \lambda}^{[t]}( \, I_t \, )$ be the maximum with respect to $I_{t+1}$, and $\omega_{ \sigma, \lambda}^{[t]}( \, I_t \,)$ be a maximizing value of $I_{t+1}$, respectively, of the function: 
\[
I_t \quad \mapsto \quad \phi_{ \sigma, \lambda}^{[t+1]}( \, I_{t}, \, I_{t+1} \, ) \, + \, \psi_{ \sigma, \lambda}^{[t+1]}( \, I_{t+1} \, )
\]
\emph{Note:} This is a finite maximization, which can be carried by considering all values of $I_{t+1}$ such that \linebreak the pair $( \, I_t, \, I_{t+1})$ satisfies Inequality (\ref{eq:It_LB_UB}). 
}
}
\tcp{FORWARD PASS: Constructs the optimal sequence of $I_t$'s}
\ForEach{ $t \in ( \, 1, \, \dots, \, T \, )$ }
{
let $I_t^* = \omega_{ \sigma, \lambda}^{[t-1]}( \, I_{t-1} \, )$
}
\Return{$I^* \define ( \, I_1^*, \, \dots, \, I_T^* \, )$}
\end{algorithm}

\section{M-step: Estimation of the Sale and Loss Distribution Parameters} \label{sec:M-step}
\vspace{0.5\baselineskip}

In this section we seek to compute values of the sale and loss distribution parameters $\sigma \in \ReSP$ and $\lambda \in [0,1]$, \linebreak given the initial inventory level $I_0 \in \ZNN$, a feasible inventory level history $I \in \ZNN^T$, the sales history $S \in \ZNN^T$, \linebreak and the replenishment history $R \in \ZNN^T$. To start, we note that in light of the arguments put forth in Section \ref{sec:E-step}, \linebreak this task is equivalent to that of maximizing function $\Phi_{ \sigma, \lambda}$ over all feasible values of the parameters $\sigma$ and $\lambda$. In turn, from Equations (\ref{eq:JLL_DBN_factorization})-(\ref{eq:function_Phi}) we recognize that the aforementioned function is the sum of functions which depend only on the parameter $\sigma$ with the sum of functions which depend only on the parameter $\lambda$. Indeed: 
\vspace{0.20\baselineskip}
\begin{equation}
\Phi_{ \sigma, \lambda}( \, I \, ) \; = \; 
\underbrace{ \left( \; \sum_{t=1}^{T} 
\log \, \Pr_{\sigma}( \, S_{t} \mid I_{t-1} \, ) \, \right) 
}_{ \displaystyle \define \Phi_{s}(\sigma) } \; + \; 
\underbrace{ \left( \; \sum_{t=1}^{T} 
\log \, \Pr_{\lambda}( \, I_t \mid I_{t-1}, \, S_t, \, R_t \, ) \, \right) 
}_{ \displaystyle \define \Phi_{\ell}(\lambda) } 
\label{eq:Phi_sigma_lambda}
\end{equation}

Therefore, computing the optimal values of the parameters $\sigma$ and $\lambda$ can be readily carried by separately computing \linebreak the maximizers of the single-argument functions $\Phi_s$ and $\Phi_{\ell}$ over their respective domains, which in turn can be \linebreak executed using any method for optimizing continuously differentiable nonlinear functions, \eg gradient ascent, \linebreak Newton's Method, BFGS, etc. 

\section{Inventory Level Estimation} \label{sec:BP}
\vspace{0.5\baselineskip}

Once the sale and loss distribution parameters have been estimated, the probability distribution over inventory levels conditional on the initial inventory and on the sales and replenishments up to period $t$, \ie 
\vspace{0.20\baselineskip}
\begin{equation}
\rho_t ( \, I_t \, ) \; \define \; 
\Pr_{ \sigma, \lambda}
( \, I_t \mid I_0, \, S_1, \, \dots, \, S_t, \, R_1, \, \dots, \, R_t \, ) \, ,
\label{eq:rho_t_def}
\end{equation}

can be easily and efficiently computed. For this purpose, we first note that $\rho_1 ( \, I_1 \, ) = \Pr_{ \sigma, \lambda} ( \, I_1 \mid I_0, \, S_1, \, R_1 \, )$ can be computed directly given the assumptions put forth in Section \ref{sec:Probabilistic_Assumptions}. Next, as noted in Ref. \cite{DeHoratius.MSOM.08}, the probability distributions associated with periods $t \geq 2$ can be sequentially computed according to the following recursion: 
\vspace{0.20\baselineskip}
\begin{equation}
\forall \, I_t \in \{ \, 0, \, \dots, \, I_t^{max} \, \} \; \colon \; 
\rho_t ( \, I_t \, ) \; = \; 
\sum_{ I_{t-1} \, = \, 0 }^{ I_{t-1}^{max} } \rho_{t-1} ( \, I_{t-1} \, ) \, 
\Pr_{ \sigma, \lambda} ( \, I_t \mid I_{t-1}, \, S_t, \, R_t \, )
\label{eq:rho_t_recursion}
\end{equation}

Lastly, once we have computed $\rho_t ( \, I_t \, )$, we can calculate the Marginal Maximum Likelihood Estimate (MMLE) of the inventory level conditional on all the information observed up to time $t$ as: 
\vspace{0.20\baselineskip}
\begin{equation}
I_t^{MMLE} \; \define \; \argmax_{ I_t \, \in \, \{ \, 0, \, \dots, \, I_t^{max} \, \} } \; 
\rho_t ( \, I_t \, )
\label{eq:conditional_MLE}
\end{equation}

\section{Experiments and Results} \label{sec:Experiments}
\vspace{0.5\baselineskip}

In this section we experimentally evaluate the performance of our EM Algorithm by means of simple simulations. \linebreak In particular, we simulate a `naive' inventory management system which computes the current inventory as the previous inventory minus current sales plus current replenishments (\ie $I_t = I_{t-1} - S_t + R_t$ for each period $t$) and follows a $( \mathcal{Q}, \, \mathcal{R} \, )$ policy. The system begins with an initial inventory level of $I_0 = 15$ units and it re-orders an amount of $\mathcal{Q} = 20$ units every time the inventory level reaches or falls below $\mathcal{R} = 10$ units. We simulate the naive system's evolution by feeding it random sales with the distributions described in Section \ref{sec:Probabilistic_Assumptions} and parameter $\sigma = 5.0$, which is sampled considering the true physical inventory. Furthermore, we setup random losses with parameter $\lambda = 0.25$, which over a time horizon of $T = 60$ periods usually causes the naive system to freeze. 

\begin{figure}[htb]
\centering
\includegraphics[ width = \columnwidth]{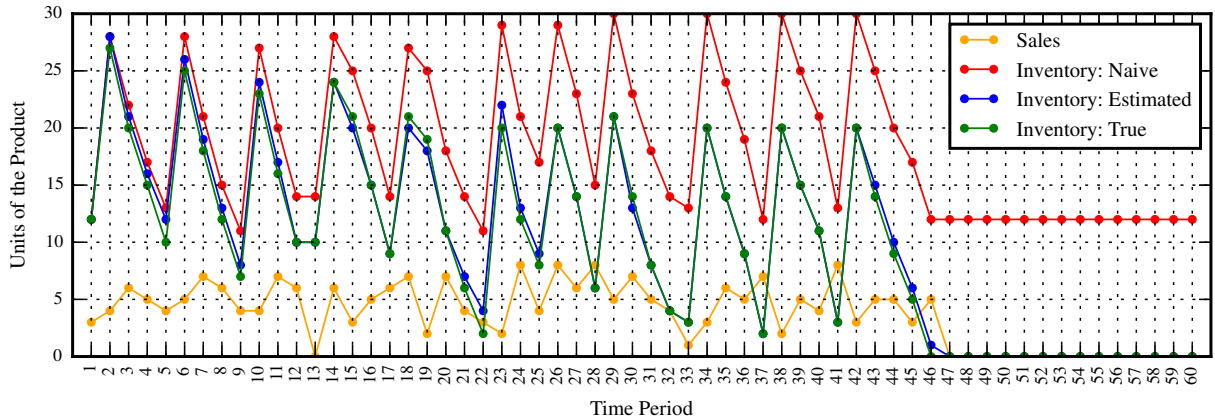}
\caption{Sales and inventory time histories for a typical simulation. Here the initial guesses where $\sigma^{[0]} = 3.72$ and $\lambda^{[0]} = 0.50$, respectively. For this example, our EM Algorithm terminated after only two iterations, with estimates $\sigma^* = 5.20$ and $\lambda^* = 0.32$. }
\label{fig:time_history_results}
\end{figure}

With this setup in place, we estimate the true physical inventory by running our EM Algorithm until the absolute changes of the estimates of $\sigma$ and $\lambda$ are within 0.01 units, respectively. Regarding the initial guesses of the parameters, \linebreak we choose $\sigma^{[0]}$ to be the average of the sales during the experiment's horizon, as it seems like a reasonable choice, and we choose $\lambda^{[0]} = 0.5$, as it is the midpoint of the interval $[0,1]$. The E-step is carried exactly as described in Section \ref{sec:E-step}, \linebreak while the M-step is carried approximately using 20 iterations of gradient ascent. Figure \ref{fig:time_history_results} shows the sale and inventory level histories for a typical simulation, where the inventory level according to the naive system is shown in red, \linebreak the MMLE of the inventory level conditional on all previous observations (see Equation (\ref{eq:conditional_MLE})) is shown in blue, and the true physical inventory level is shown in green. 

Noteworthily, for most of the examples we simulated our EM Algorithm terminated after less than five iterations. 

\section{Future Research} \label{sec:Future_Research}
\vspace{0.5\baselineskip}

There are several lines of future research which can stem from the work presented in this paper. For instance, \linebreak the performance of our EM Algorithm should be evaluated on real data and compared to the performance of similar \linebreak methods proposed in the literature. This would be interesting because in the real world the true physical inventory \linebreak levels are not known. Another avenue of research is the extension of our model to one that can optimally decide on product re-orders and inventory inspections. This would be challenging because it leads to a Partially-observable Markov Decision Process (POMDP) formulation, and its is well-known that even approximating the optimal policies for POMDPs can be, in general, computationally intractable. 

\section*{Acknowledgements}
\vspace{0.5\baselineskip}

The authors would like to thank the Board of Directors of Tiendas Industriales Asociadas Sociedad An\'onima \linebreak (TIA S.A.), a leading grocery retailer in Ecuador, for authorizing their company to provide the authors with time history data of sales, replenishments and inventory levels for hundreds of their products and several of their stores. Explorative analysis of this data lead to the discovery of products which seem to have suffered from stock freezing, which in turn lead to the design of our DBN model and our EM algorithm. 

\section*{References}
\vspace{0.5\baselineskip}

\bibliographystyle{plain}
\renewcommand{\section}[2]{}%
\bibliography{Reyes.Abad.ISERC16.Refs}

\end{document}

%% file: Fig_DBN_Model.eps_tex
\begingroup%
  \makeatletter%
  \providecommand\color[2][]{%
    \errmessage{(Inkscape) Color is used for the text in Inkscape, but the package 'color.sty' is not loaded}%
    \renewcommand\color[2][]{}%
  }%
  \providecommand\transparent[1]{%
    \errmessage{(Inkscape) Transparency is used (non-zero) for the text in Inkscape, but the package 'transparent.sty' is not loaded}%
    \renewcommand\transparent[1]{}%
  }%
  \providecommand\rotatebox[2]{#2}%
  \ifx\svgwidth\undefined%
    \setlength{\unitlength}{320.55bp}%
    \ifx\svgscale\undefined%
      \relax%
    \else%
      \setlength{\unitlength}{\unitlength * \real{\svgscale}}%
    \fi%
  \else%
    \setlength{\unitlength}{\svgwidth}%
  \fi%
  \global\let\svgwidth\undefined%
  \global\let\svgscale\undefined%
  \makeatother%
  \begin{picture}(1,0.30120106)%
    \put(0,0){\includegraphics[width=\unitlength]{Fig_DBN_Model.eps}}%
    \put(0.73958821,0.14064967){\color[rgb]{0,0,0}\makebox(0,0)[lb]{\smash{$I_{T-1}$}}}%
    \put(0.405163,0.14064966){\color[rgb]{0,0,0}\makebox(0,0)[lb]{\smash{$I_2$}}}%
    \put(0.54242707,0.14564111){\color[rgb]{0,0,0}\makebox(0,0)[lb]{\smash{$\mathbf{\cdots}$}}}%
    \put(0.23046327,0.14064967){\color[rgb]{0,0,0}\makebox(0,0)[lb]{\smash{$I_1$}}}%
    \put(0.05576353,0.14064967){\color[rgb]{0,0,0}\makebox(0,0)[lb]{\smash{$I_0$}}}%
    \put(0.92177507,0.14064967){\color[rgb]{0,0,0}\makebox(0,0)[lb]{\smash{$I_T$}}}%
    \put(0.64724693,0.05329981){\color[rgb]{0,0,0}\makebox(0,0)[lb]{\smash{$S_{T-1}$}}}%
    \put(0.31282171,0.05329981){\color[rgb]{0,0,0}\makebox(0,0)[lb]{\smash{$S_2$}}}%
    \put(0.13812198,0.05329981){\color[rgb]{0,0,0}\makebox(0,0)[lb]{\smash{$S_1$}}}%
    \put(0.83192951,0.05329981){\color[rgb]{0,0,0}\makebox(0,0)[lb]{\smash{$S_T$}}}%
    \put(0.64724693,0.24047809){\color[rgb]{0,0,0}\makebox(0,0)[lb]{\smash{$R_{T-1}$}}}%
    \put(0.31282171,0.24047809){\color[rgb]{0,0,0}\makebox(0,0)[lb]{\smash{$R_2$}}}%
    \put(0.13812198,0.24047809){\color[rgb]{0,0,0}\makebox(0,0)[lb]{\smash{$R_1$}}}%
    \put(0.83192951,0.24047809){\color[rgb]{0,0,0}\makebox(0,0)[lb]{\smash{$R_T$}}}%
    \put(0.48752145,0.05080409){\color[rgb]{0,0,0}\makebox(0,0)[lb]{\smash{$\mathbf{\cdots}$}}}%
  \end{picture}%
\endgroup%